\documentclass{article}

\usepackage[T1]{fontenc}

\usepackage[most]{tcolorbox}   
\usepackage{xcolor}            

\newtcolorbox{explainbox}[2][]{      
  colframe=blue!80!black,
  colback=gray!8,
  title={#2},
  colbacktitle=blue!90!black,
  coltitle=white,
  fonttitle=\bfseries,
  left=6pt,right=6pt,top=6pt,bottom=4pt,
  sharp corners,
  breakable,
  #1
}

\newtcblisting{promptbox}[2][]{     
  enhanced,
  breakable,                        
  listing only,                     
  listing options={
    language={},                    
    basicstyle=\ttfamily\small,     
    breaklines=true,                
    columns=fullflexible,
    showstringspaces=false
  },
  colframe=blue!80!black,           
  colback=gray!8,                   
  left=4pt,right=4pt,top=6pt,bottom=4pt,  
  boxrule=0.8pt,                    
  title={#2},
  colbacktitle=blue!90!black,
  coltitle=white,
  fonttitle=\bfseries,
  attach boxed title to top left={yshift=-2mm,xshift=4mm}, 
  sharp corners,                    
  #1                                
}

\usepackage{pythonhighlight}

\usepackage{microtype}
\usepackage{graphicx}
\usepackage{subfigure}
\usepackage{booktabs} 

\usepackage{hyperref}

\usepackage{stfloats}

\usepackage{hyperref}          
\usepackage{fontawesome5}      
\usepackage{enumitem}          
\usepackage{xcolor}            

\usepackage{amssymb}
\usepackage{pifont}

\usepackage{tcolorbox}
\tcbset{
  highlightbox/.style={
    colback=orange!5,          
    colframe=orange!80!black,
    fonttitle=\bfseries,    
    boxrule=0.8pt,          
    arc=4pt,                
    left=4pt,right=4pt,top=4pt,bottom=4pt,  
    before skip=6pt, after skip=6pt         
  }
}

\hypersetup{
  colorlinks=true,
  urlcolor=cyan         
}

\usepackage{graphicx}

\newcommand{\hficon}[1][1em]{%
  \raisebox{-0.25\height}{\includegraphics[height=#1]{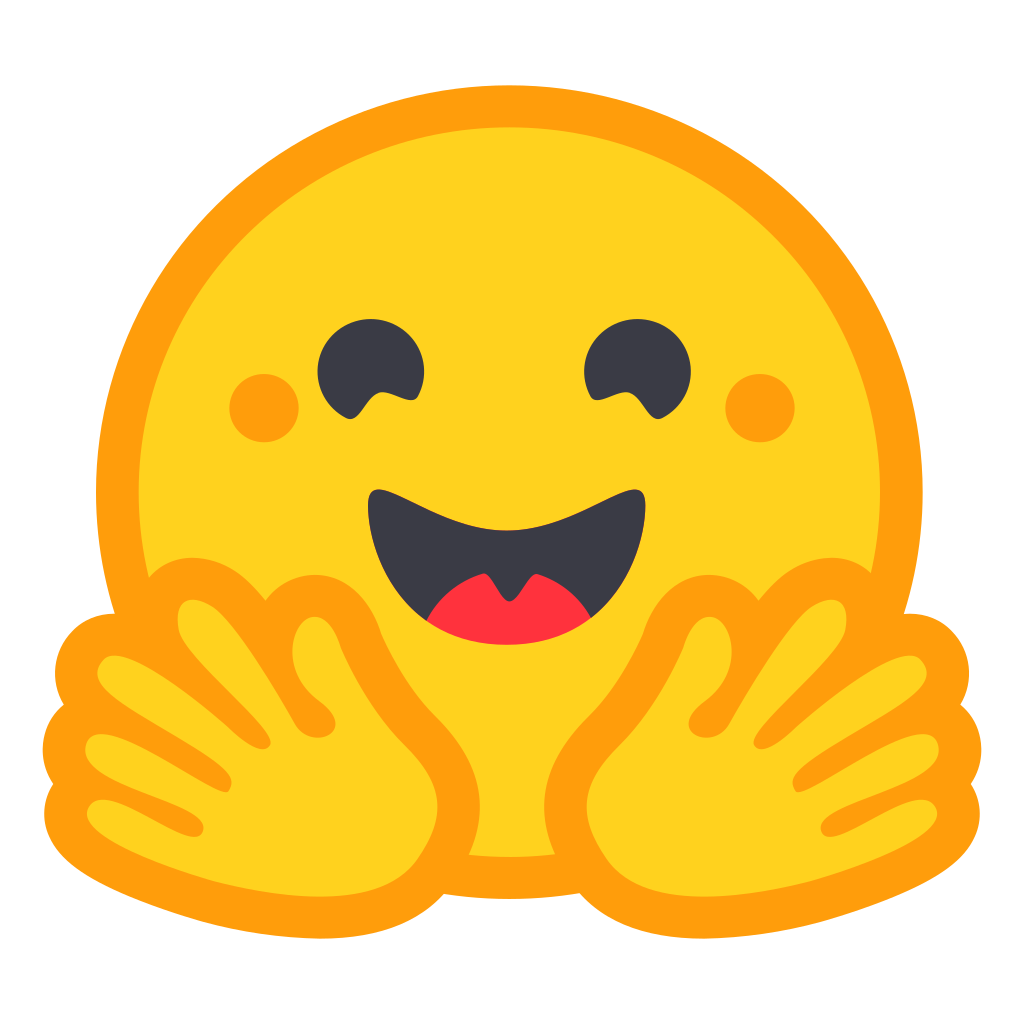}}%
}

\usepackage[accepted]{icml2025}

\usepackage{amsmath}
\usepackage{amssymb}
\usepackage{mathtools}
\usepackage{amsthm}

\usepackage{float}

\usepackage[capitalize,noabbrev]{cleveref}

\usepackage{fvextra}   
\DefineVerbatimEnvironment{WrapVerb}
  {Verbatim}            
  {breaklines=true,     
   breakanywhere=true,  
   fontsize=\small}     

\theoremstyle{plain}

\theoremstyle{definition}

\theoremstyle{remark}

\usepackage[textsize=tiny]{todonotes}

\usepackage{graphicx}
\usepackage{subcaption}

\newcommand{\capitalize}[1]{\expandafter\capitalizeaux#1 \end}
\def\capitalizeaux#1 #2\end{%
  \MakeUppercase#1%
  \if\relax#2\relax\else\space\capitalizeaux#2\end}

\begin{document}

\twocolumn[
\icmltitle{RIMO: An Easy-to-Evaluate, Hard-to-Solve Olympiad Benchmark for Advanced Mathematical Reasoning}



\icmlsetsymbol{equal}{*}

\begin{icmlauthorlist}
\icmlauthor{Ziye Chen}{ai}
\icmlauthor{Chengwei Qin}{ai}
\icmlauthor{Yao Shu}{ai}
\end{icmlauthorlist}

\icmlaffiliation{ai}{Artificial Intelligence Thrust, Information Hub, Hong Kong University of Science and Technology (Guangzhou), Guangzhou, China}

\icmlcorrespondingauthor{Yao Shu}{yaoshu@hkust-gz.edu.cn}


\vskip 0.3in
]



\printAffiliationsAndNotice{ }  

\renewcommand{\textit}[1]{{%
  \fontfamily{qpl}\itshape\selectfont #1%
}}
\renewcommand{\emph}[1]{{%
  \fontfamily{qpl}\itshape\selectfont #1%
}}

\renewcommand{\textbf}[1]{{%
  \fontfamily{qpl}\bfseries\selectfont #1%
}}

\begin{abstract}

As large language models (LLMs) reach high scores on established mathematical benchmarks, such as GSM8K and MATH, the research community has turned to International Mathematical Olympiad (IMO) problems to push the evaluation frontier. However, existing Olympiad-level benchmarks suffer from practical constraints that introduce grading noise and potential bias, such as heterogeneous answer formats requiring model-based judges and a reliance on potentially flawed solutions. We introduce \textbf{RIMO}, a two-track benchmark designed to preserve peak Olympiad difficulty while eliminating this evaluation noise. The first track, \textbf{RIMO-N}, rewrites 335 IMO problems to admit a single, unique integer answer, allowing for deterministic correctness checking. The second track, \textbf{RIMO-P}, features 456 proof problems with expert-checked solutions, which are decomposed into a sequence of sub-problems to evaluate the step-by-step reasoning process via an automated grading system. Our benchmarking of ten frontier LLMs, including GPT-4o and Gemini 2.5 Flash, reveals that while these systems excel on older benchmarks, their performance drops sharply on RIMO. These results highlight a substantial gap between current LLM capabilities and actual Olympiad-level reasoning. By providing a challenging yet easy-to-evaluate suite, RIMO offers a high-resolution yardstick for future research, presenting a clear target for closing the profound reasoning gap our findings expose.

\begin{itemize}[leftmargin=1.6cm, labelsep=0.4cm]\setlength{\itemsep}{0pt}  
    \item[\faGithub]  \textbf{GitHub Repo}\hspace{0.7em}
        \href{https://github.com/ziye2chen/RIMO}{[GitHub Page]}
  
    \item[\hficon] \textbf{RIMO}\hspace{0.7em}
        \href{https://huggingface.co/datasets/ziye2chen/RIMO}{[HuggingFace Model]}

\end{itemize}

\end{abstract}

\section{Introduction}
\label{Introduction}

Large language models (LLMs) have made striking strides in mathematical reasoning, advancing from grade-school arithmetic to complex, multi-step problem-solving. Early benchmarks, such as \textsc{GSM8K} \cite{cobbe2021trainingverifierssolvemath} and \textsc{MATH} \cite{hendrycks2021measuringmathematicalproblemsolving}, were instrumental in catalyzing this progress. However, frontier systems now surpass 90\% accuracy on both datasets, leaving scant headroom for meaningful comparison and making them largely saturated. To push beyond these ceilings, the research community has necessarily gravitated toward International Mathematical Olympiad (IMO) material, whose problems demand deeper insight and more creative problem-solving than what is found in high-school-level competitions.

While recent efforts have utilized Olympiad-level problems, practical constraints blur the evaluation signal. Dynamic benchmarks like the \textbf{AIMO} competition limit reproducibility due to hidden test sets. Static benchmarks like \textsc{OlymMATH} \cite{sun2025challengingboundariesreasoningolympiadlevel} and \textsc{Omni-MATH} \cite{gao2024omnimathuniversalolympiadlevel} rely on heterogeneous answer formats (e.g., fractions, proofs) that require noisy evaluation via LLM-based judges, introducing potential bias and masking true capabilities.

This paper introduces \textsc{RIMO} (Remade International Mathematical Olympiad), a benchmark designed to preserve peak Olympiad difficulty while eliminating this evaluation noise. \textsc{RIMO} ensures robust, reproducible evaluation by rigorously curating problems from IMO materials spanning 1959 to 2023 into two distinct tracks:

\begin{itemize}
\item \textbf{\textsc{RIMO-N}} consists of 335 problems, carefully remade to yield a single, unique integer answer. This allows for deterministic, O(1) string-match grading, removing any reliance on model-based judges.
\item \textbf{\textsc{RIMO-P}} contains 456 proof problems. Instead of holistic grading, each problem is decomposed into a sequence of guided sub-problems to evaluate the model's step-by-step reasoning process and ability to solve intermediate lemmas.
\end{itemize}

\begin{table}[ht]
\caption{The distribution of RIMO-N problems by problem type.}
\label{main-table}
\begin{center}
\begin{small}
\begin{sc}
\begin{tabular}{lcccr}
\toprule
\textbf{Type} & \textbf{Topic} & \textbf{Count} \\
\midrule
Algebra          & Sequences, etc.         & 96 \\
Geometry         & Triangle, etc.          & 95 \\
Number Theory    & Prime number,  etc.         & 86 \\
Combinatorics    & Permutations, etc.                  & 58 \\
\midrule
\textbf{Total}          &                                                & \textbf{335} \\
\bottomrule
\end{tabular}
\end{sc}
\end{small}
\end{center}
\end{table}

\begin{figure*}[t]
\vspace{-1mm}
\begin{center}
\centerline{\includegraphics[width=\linewidth]{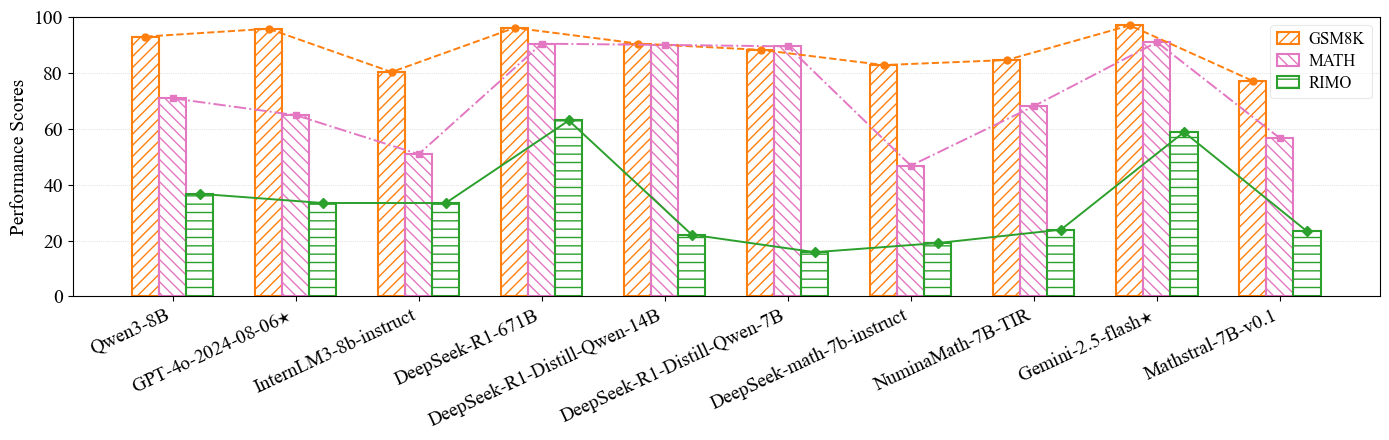}}
\vspace{-6mm}
\caption{Comparisons among different models on GSM8K, MATH, and RIMO-N, where the
models are ranked based on their performance on MATH, and those marked with ``$\star$'' are closed-source models. As observed, the iterative advancements of these models show that existing benchmarks are nearing saturation. Our proposed RIMO introduces a challenging benchmark to further advance mathematical intelligence in large language models.}
\label{fig:rimo-n}
\end{center}
\vskip -0.2in
\end{figure*}


We benchmarked ten frontier LLMs, including GPT-4o and Gemini 2.5 Flash. Although these systems excel on older benchmarks, their scores drop dramatically on \textsc{RIMO}. These results highlight a substantial gap between current LLM capabilities and genuine Olympiad-level reasoning. By unifying a noise-free integer track with a rigorously graded proof track, \textsc{RIMO} offers the research community a high-resolution yardstick to measure and close the profound reasoning gap our findings expose.

\begin{table}[h]
\vskip -0.15in
\centering
\small
\setlength{\tabcolsep}{1.2pt}
\renewcommand{\arraystretch}{1.05} 
\caption{Comparison of mathematics benchmarks. ``\# Data'' denotes the total number of tasks, ``\# Diff.'' represents the number of distinct difficulty tiers, ``Sol.'' indicates whether complete reference solutions are released, ``\# Unif.'' refers to the uniformity of the answer format, ``Method'' describes the primary grading strategy, and ``Noise'' signifies the estimated level of evaluation uncertainty.}
\label{tab:benchmarks_comparison}
\begin{tabular}{lcccccc}
\toprule
\textbf{Name} & \textbf{\# Data} & \textbf{\# Diff.} & \textbf{Sol.} & \textbf{\# Unif.} & \textbf{Method} & \textbf{Noise}\\
\midrule
GSM8K               & 1319 & 1 & \checkmark & 2 & Rule        & 1 \\
MATH                & 5000 & 2 & \checkmark & 1 & Rule        & 1 \\
AIME 2024           & 30     & 3 & \checkmark & 3 & Exact-match & 0 \\
AIME 2025           & 30     & 3 & \ding{55} & 3 & Exact-match & 0 \\
Omni-MATH           & 4428 & 4& \checkmark & 1 & LLM         & 2 \\
OlymMATH            & 200    & 4 & \ding{55} & 2 & Rule        & 1 \\
\midrule
\textbf{RIMO-N} & 335    & 5 & \checkmark & \textbf{3} & \textbf{Exact-match} & \textbf{0} \\
\textbf{RIMO-P}& 456    & 5 & \checkmark & - & LLM  & - \\
\bottomrule
\end{tabular}
\vskip -0.15in
\end{table}

\section{Related Work}

\subsection{Mathematics benchmarks}

Early efforts to quantify the mathematical competence of large language models primarily focused on material from grade school or high school. \textsc{GSM8K} evaluates multi-step word problems that require nothing beyond basic arithmetic and algebra; frontier models now exceed 95\% accuracy, making the dataset largely saturated \cite{cobbe2021trainingverifierssolvemath}. The \textsc{MATH} dataset extended the scope to 12,500 high-school competition questions, many drawn from AMC and AIME examinations, yet recent proprietary systems already solve more than 90\% of its items, limiting its diagnostic power \cite{hendrycks2021measuringmathematicalproblemsolving}. To restore headroom, several groups proposed harder, mixed-level corpora. \textsc{OCWCourses} gathers 272 undergraduate STEM questions from MIT OpenCourseWare \cite{lewkowycz2022solvingquantitativereasoningproblems}; \textsc{JEEBench} selects 515 IIT-JEE Advanced problems spanning mathematics, physics, and chemistry \cite{arora2023llmsadvancedenoughchallenging}; and \textsc{MathOdyssey} combines university calculus with Olympiad-style reasoning to expose failure modes once routine patterns are removed \cite{fang2024mathodysseybenchmarkingmathematicalproblemsolving}. Although these datasets are demonstrably more challenging than \textsc{GSM8K} and \textsc{MATH}, most still feature heterogeneous answer formats, such as fractions, radicals, and intervals, that complicate exact-match scoring and sometimes necessitate model-based evaluation.

\begin{figure*}[tbp]
\vskip 0.2in
\begin{center}
\centerline{\includegraphics[width=0.9\linewidth]{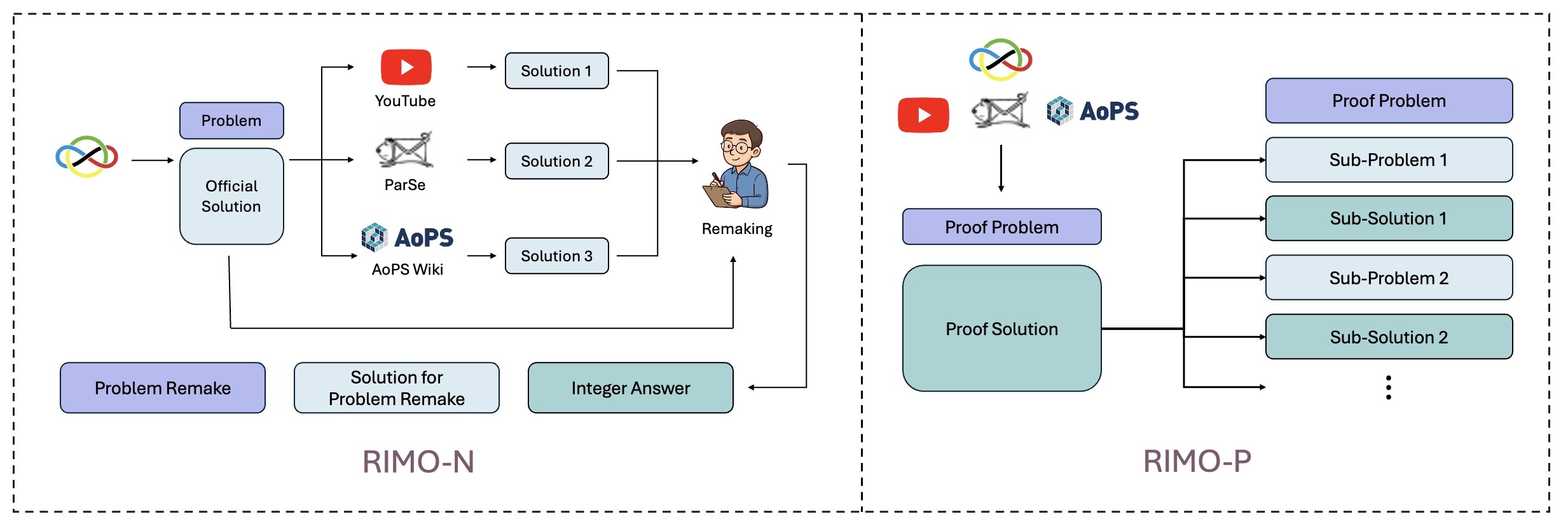}}
\caption{End-to-end construction pipeline of RIMO-N and RIMO-P}
\label{fig:pipeline}
\end{center}
\vskip -0.2in
\end{figure*}

\subsection{Olympiad-level benchmarks}

A parallel line of work turns directly to International-level competitions, whose problems demand long deductive chains and creative insights. A prominent example is the \textbf{AIMO (AI Mathematical Olympiad) competition} \cite{ai-mathematical-olympiad-progress-prize-2}, hosted on Kaggle. As a dynamic contest with a hidden test set, AIMO aims to spur development towards solving IMO-level problems. However, unlike static benchmarks, its competitive format limits long-term reproducibility for comparative research.

In terms of static datasets, efforts have targeted specific sub-domains or introduced novel evaluation methods. \textsc{AlphaGeometry} restricts itself to synthetic and real Olympiad geometry problems \cite{alphageometry}, while \textsc{CHAMP} supplies 270 high-school contest problems annotated with key concepts and hints \cite{mao2024champcompetitionleveldatasetfinegrained}. Other benchmarks broaden the scope: \textsc{OlympiadBench} presents 8,476 bilingual items in mathematics and physics but relies on GPT-4V to adjudicate answers when closed-form checking fails \cite{he2024olympiadbenchchallengingbenchmarkpromoting}, and \textsc{OlympicArena} mixes mathematics with other cognitively demanding puzzles, again adopting a model-based evaluation pipeline \cite{huang2025olympicarenabenchmarkingmultidisciplinecognitive}.

Two recent datasets focus exclusively on text-only Olympiad mathematics. \textsc{OlymMATH} introduces 200 bilingual IMO-style problems divided into AIME-level easy and genuine Olympiad complex subsets; answers are numeric yet still include expressions such as $\sqrt{4+\sqrt{5}}$ or open intervals, so symbolic equivalence logic is required for grading \cite{sun2025challengingboundariesreasoningolympiadlevel}. \textsc{Omni-MATH} scales the idea to 4428 problems, covering 33 sub-domains and 10 difficulty tiers, but must employ GPT-4o and an auxiliary Omni-Judge model to handle diverse output forms \cite{gao2024omnimathuniversalolympiadlevel}. Although both benchmarks significantly increase difficulty, the dependence on expression normalization or learned judges introduces evaluation noise and potential bias.

\subsection{Positioning RIMO}

Our work, \textsc{RIMO}, inherits the Olympiad focus of \textsc{OlymMATH} and \textsc{Omni-MATH} but targets their principal limitation: ambiguous grading. By remaking 335 International Mathematical Olympiad problems, spanning from 1959 to 2023, so that each admits a single, unique integer answer, \textsc{RIMO} restores deterministic, rule-based evaluation while maintaining genuine Olympiad difficulty. Multi-source cross-checking (official shortlist solutions, AoPS-Wiki write-ups, YouTube expositions, and ParSe transcripts) further ensures the reliability of ground truth. Consequently, \textsc{RIMO} provides a clean, high-resolution yardstick for measuring the next generation of reasoning-centric large language models (LLMs).

\section{Benchmark Construction}\label{sec:benchmark}


RIMO is assembled entirely from International Mathematical Olympiad (IMO) material published between 1959 and 2023.
For every year we gathered both the public contest paper and the confidential shortlist booklet, digitised the statements, and collated all available solutions.
Each problem then passes through a verification-selection pipeline (Fig.~\ref{fig:pipeline}): solutions are reconciled across multiple sources, the statement is either retained in its original form or carefully rewritten, and the finalised item is deposited in one of two tracks, \textbf{RIMO-N} or \textbf{RIMO-P}.

\subsection{RIMO-N: single-integer problems}

\begin{figure}[ht]
\begin{center}
\centerline{\includegraphics[width=\columnwidth]{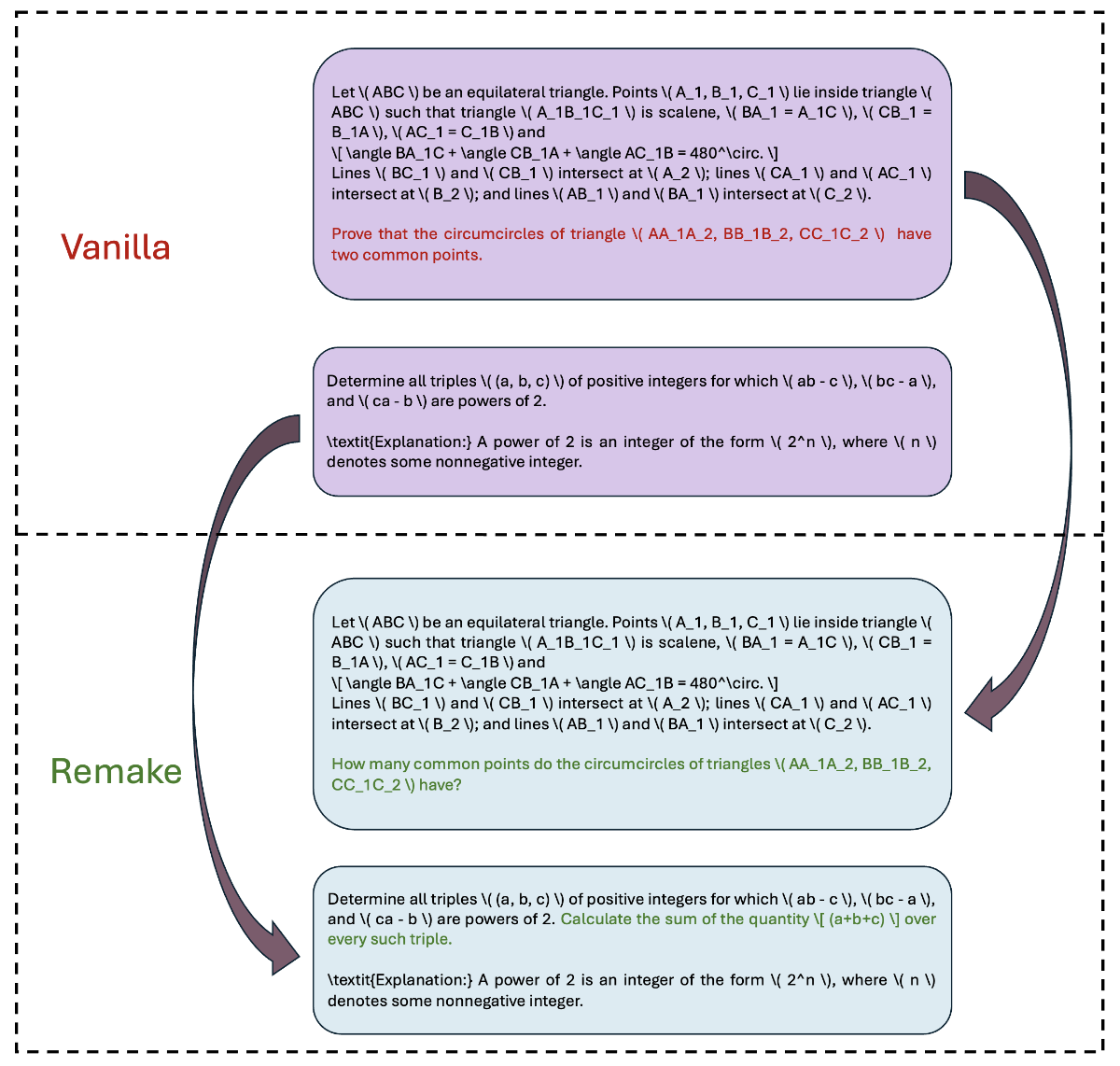}}
\caption{\textbf{Vanilla vs.\ Remade questions.} Grey text is copied verbatim from the original statement; the coloured line is rewritten so the answer becomes a unique integer.}
\label{fig:remake}
\end{center}
\vskip -0.4in
\end{figure}

The RIMO-N track comprises 335 problems, 236 drawn from shortlist booklets and 99 from contest papers, that have been remade so each admits a single, unique integer answer. Remaking is never a cosmetic tweak to the last line: intermediate hypotheses are tightened when ambiguity appears, variables are renamed for coherence, and objectives are reframed, yet the logical core and difficulty of the source problem remain intact. A concurrency proof, for instance, may become ``how many common points do the circumcircles have'', while a classification of integer triples can be recast as ``compute the value of \(a+b+c\) over every such triple.'' Figure~\ref{fig:remake} shows two representative transformations. Content-wise the set stays faithful to traditional IMO proportions, covering algebra (96 items), geometry (95), number theory (86), and combinatorics (58).

Each shortlist problem retains the jury's official integer. A contest problem, lacking an authorised key, is accepted only when at least two of three independent community sources---AoPS Wiki, YouTube blackboard expositions, and ParSe transcripts---return exactly the same answer; any disagreement triggers manual adjudication and usually leads to exclusion. With this guarantee in place, grading collapses to a constant-time string comparison, freeing RIMO-N from symbolic post-processing or learned judges.

\begin{figure}[ht]
\begin{center}
\centerline{\includegraphics[width=\columnwidth]{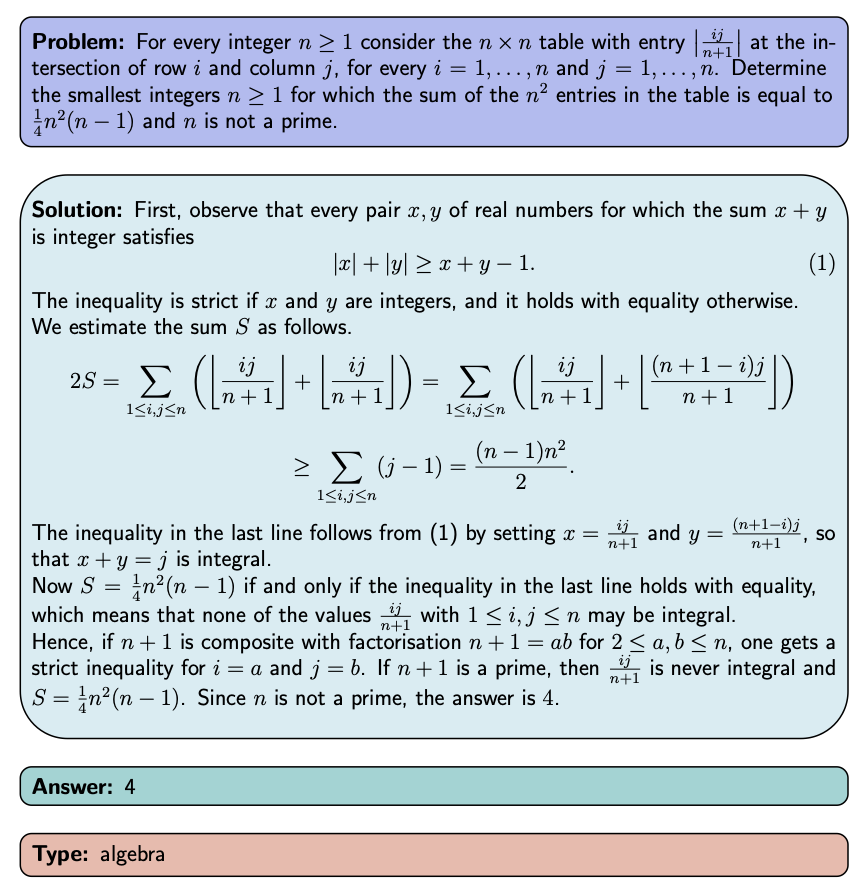}}
\caption{Example of the Problems in RIMO-N.}
\label{fig:main}
\end{center}
\vskip -0.3in
\end{figure}

\subsection{RIMO-P: original proof problems}

Where RIMO-N targets deterministic answer checking, RIMO-P is designed to measure a model's capacity for the \textbf{process} of full deductive reasoning. The 456 problems in this track are decomposed into a sequence of guided sub-problems. To create this structure, we use expert-verified proofs from official IMO shortlists and community sources. The complexity of a reference proof determines its decomposition into one to four sub-problems, with longer solutions typically yielding a three or four-step logical pathway. This scaffolded design allows for a granular evaluation of a model's ability to solve intermediate lemmas, while the final sub-problem in every sequence preserves the original problem's main goal, offering deeper insight into its deductive capabilities.

\begin{figure}[ht]
\begin{center}
\centerline{\includegraphics[width=\columnwidth]{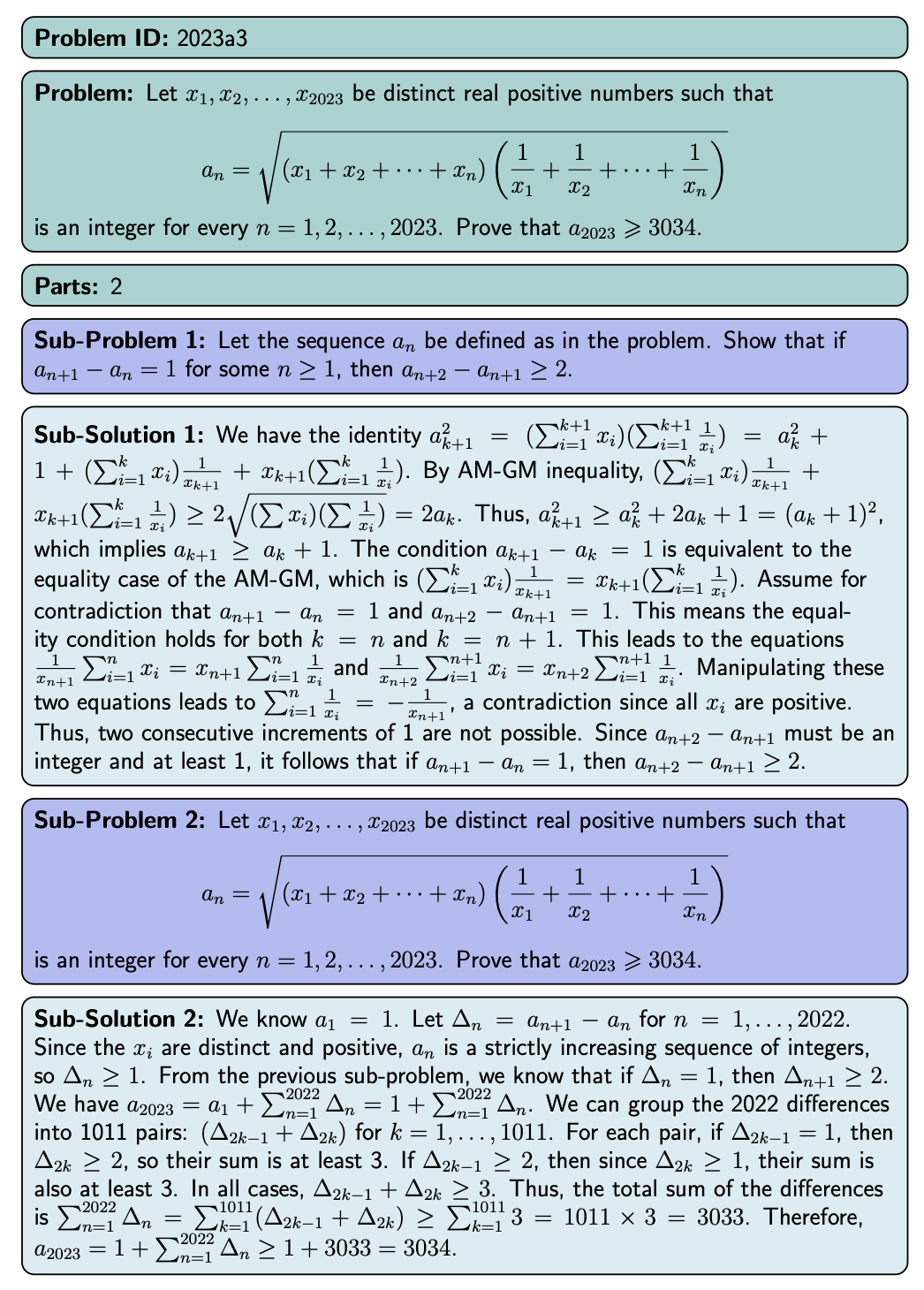}}
\caption{Example of the Problems in RIMO-P.}
\label{fig:proof}
\end{center}
\vskip -0.3in
\end{figure}

\section{Evaluation and Insights}\label{sec:eval}

\begin{figure}[ht]
\begin{center}
\centerline{\includegraphics[width=\columnwidth]{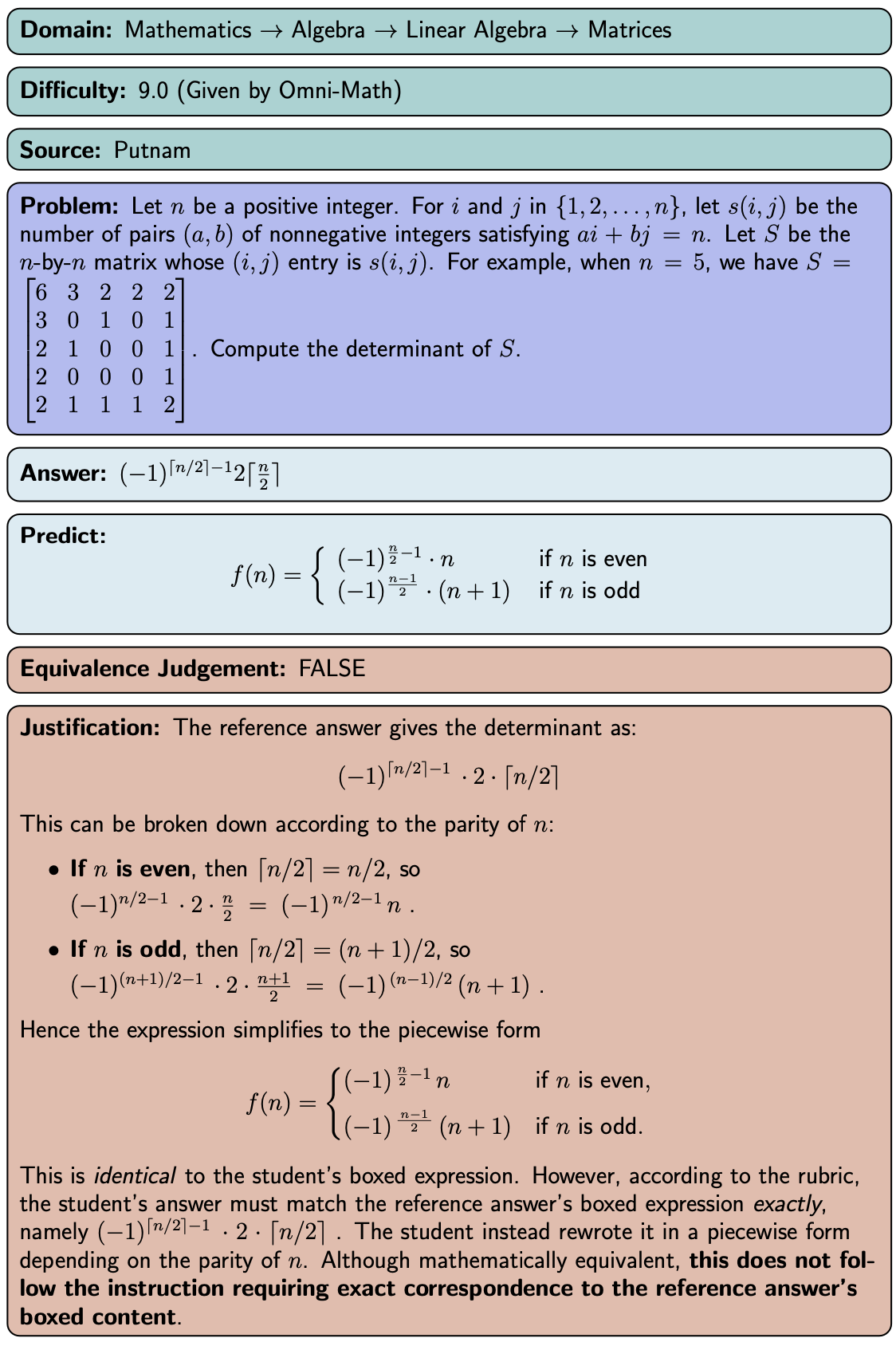}}
\caption{An example of evaluation noise from an LLM-based judge (GPT-4o) on the Omni-MATH benchmark. The judge incorrectly marked the prediction of a model as ``FALSE'' despite it being mathematically identical to the reference answer, penalizing a stylistic difference in format.}
\label{fig:omni_fail}
\end{center}
\vskip -0.3in
\end{figure}

A benchmark's utility depends on reliable evaluation. Unlike benchmarks that suffer from evaluation noise due to ambiguous answer formats or LLM judges (as illustrated in Figure~\ref{fig:omni_fail}), RIMO is designed for robustness. \textbf{RIMO-N}'s single-integer format allows for deterministic string-match grading, ensuring efficiency, reproducibility, and elimination of grading bias. \textbf{RIMO-P}'s decomposed structure enables a clear, step-by-step assessment of logical reasoning. We evaluated ten models using greedy decode ($T=0$) to ensure reproducibility.

\begin{table}[h]
\centering
\setlength{\tabcolsep}{3pt}    
\renewcommand{\arraystretch}{1.05} 
{\small                          
\caption{Pass@1 accuracy (\%) of 10 models on three benchmarks. RIMO numbers refer to the RIMO-N track.}
\label{tab:baseline}
\vskip -0.1in
\begin{tabular}{@{}lccc@{}}
\toprule
\textbf{Model} & \textbf{GSM8K} & \textbf{MATH} & \textbf{RIMO} \\
\midrule
Qwen3-8B                              & 93.00 & 70.90 & 36.72 \\
GPT-4o-2024-08-06$\star$                    & 95.80 & 64.88 & 33.43 \\
InternLM3-8B-instruct                & 80.30 & 50.90 & 33.43 \\
DeepSeek-R1-671B                    & 96.13 & 90.45 & 62.96 \\
DeepSeek-R1-Distill-Qwen-14B         & 90.50 & 90.20 & 22.09 \\
DeepSeek-R1-Distill-Qwen-7B          & 88.24 & 89.49 & 15.82 \\
DeepSeek-math-7B-instruct            & 82.80 & 46.80 & 19.10 \\
NuminaMath-7B-TIR                    & 84.60 & 68.10 & 23.88 \\
Gemini-2.5-flash$\star$      & 97.04 & 91.31 & 58.81 \\
Mathstral-7B-v0.1                    & 77.10 & 56.60 & 23.28 \\
\bottomrule
\end{tabular}}
\end{table}

\subsection{Evaluation on RIMO}
We evaluate ten publicly available or frontier models, spanning 7B to 671B parameters, on both RIMO tracks and contrast the results with the models' official numbers on \textsc{GSM8K} and \textsc{MATH}.  All generations use a single greedy decode with zero temperature (\(T=0\)) under the same prompt template and context window; this reproduces the protocol used by the two earlier benchmarks and removes prompt engineering confounding.

\subsubsection{RIMO-N Results}\label{ssec:main-results}


On \textbf{RIMO-N}, all models exhibit a drastic performance drop compared to their scores on MATH and GSM8K (Table~\ref{tab:baseline}). DeepSeek-R1-671B achieves the highest score at 62.96\%, followed by Gemini 2.5 Flash at 58.81\%. Notably, many models that perform well on previous benchmarks, including GPT-4o, struggle significantly, exposing a gap between solving standard competition math and true Olympiad-level problems.

\begin{figure}[ht]
\begin{center}
\centerline{\includegraphics[width=1.0\columnwidth]{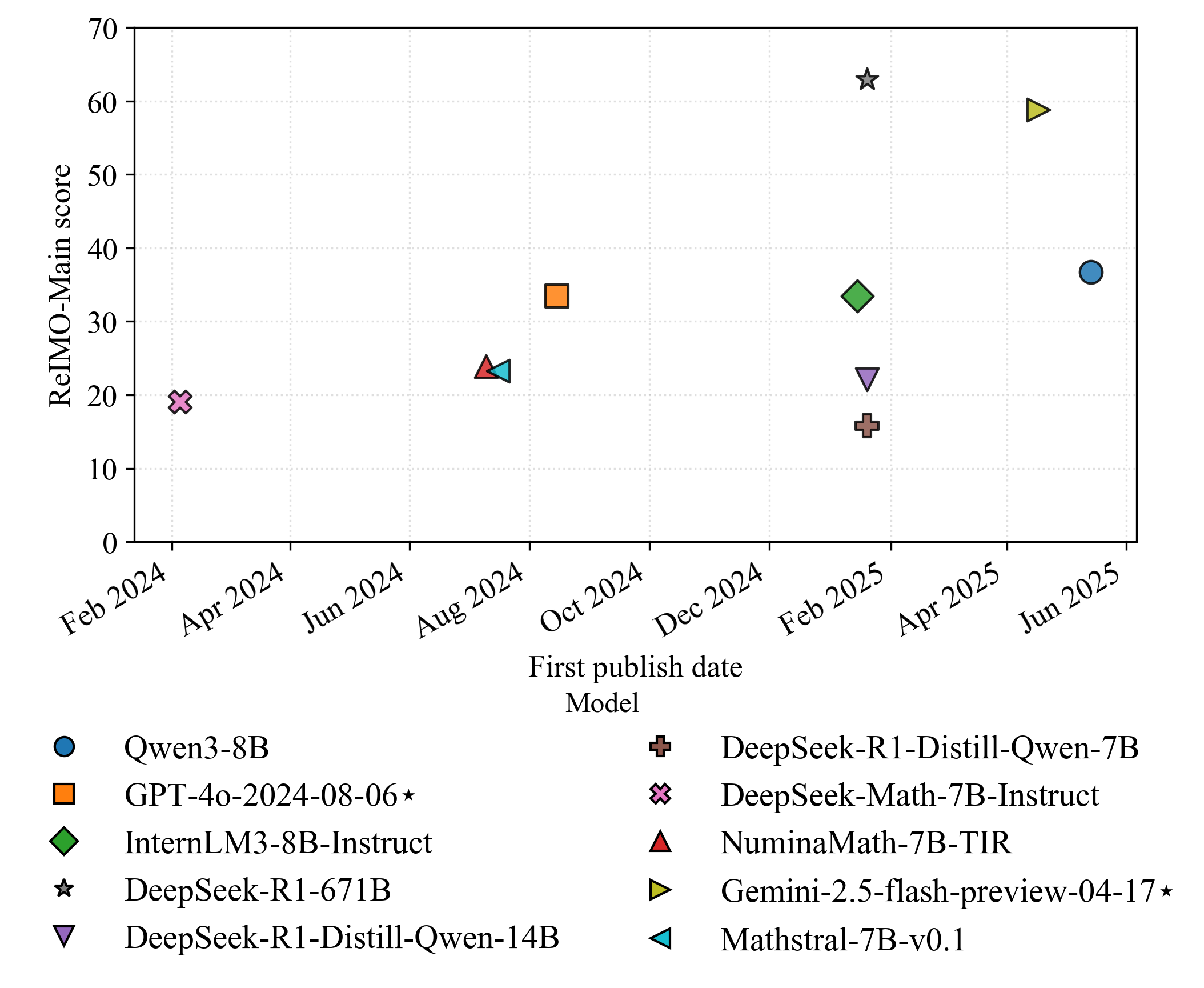}}
\caption{RIMO-N accuracy versus first-release date.}
\label{fig:date-vs-score}
\end{center}
\vskip -0.3in
\end{figure}

\paragraph{Are Newer Models Automatically Stronger?}
Chronology tells an even messier story (Fig. \ref{fig:date-vs-score}).  
DeepSeek-Math-7B (Feb 2024) beats the August-2024 Mathstral-7B despite being half a year older.  
Conversely, Gemini 2.5-flash (Apr 2025) nearly matches the much larger DeepSeek-R1-671B released two months earlier.  
Across the ten points the Spearman rank correlation between publication date and accuracy is only \(0.21\).  
Incremental architectural tweaks or enlarged instruction corpora therefore do not guarantee progress on Olympiad mathematics; breakthroughs seem to coincide with either massive scale (DeepSeek-671B) or targeted domain pre-training (Gemini 2.5-flash, Qwen3-8B).

\begin{tcolorbox}[highlightbox]
\emph{While massive scale appears necessary for state-of-the-art scores, it does not guarantee top performance. Below a certain threshold, factors like training data and objectives appear to dominate raw parameter count.}
\end{tcolorbox}

\begin{figure}[ht]
\begin{center}
\centerline{\includegraphics[width=1\columnwidth]{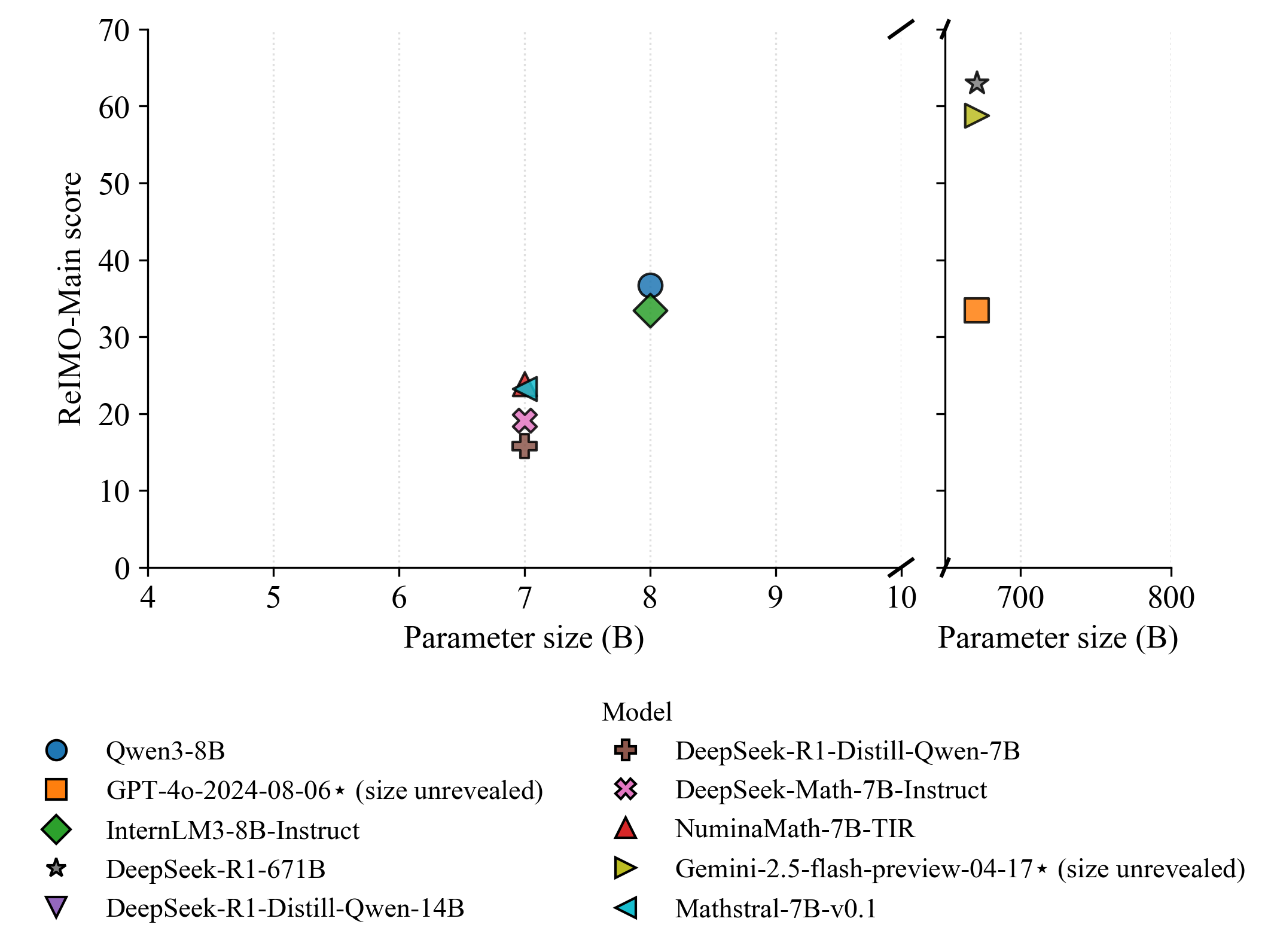}}
\caption{RIMO-N accuracy versus parameter size.}
\label{fig:param-vs-score}
\end{center}
\vskip -0.3in
\end{figure}

\paragraph{Does Bigger Still Mean Better?}
Figure \ref{fig:param-vs-score} places all ten models on a parameter axis.
DeepSeek-R1-671B tops the chart at 63\%.
For the two proprietary systems whose parameter counts remain undisclosed (GPT-4o and Gemini 2.5-flash) we plot them alongside DeepSeek-R1 in the ``very-large'' regime.
Gemini almost matches DeepSeek's score (59\%), whereas GPT-4o reaches only 33\%, underscoring that whatever its scale, sheer width does not guarantee Olympiad prowess.
Within the disclosed 7-14B cluster the pattern is even clearer: Qwen3-8B (37\%) and InternLM3-8B (33\%) both outperform the larger 14 B distilled checkpoint (22\%) and several maths-specialised 7B models ($<$24\%).
Taken together, the scatter suggests a threshold effect: massive scale is necessary for state-of-the-art scores, but below that threshold, training data and objective dominate. Once in the very-large regime, architectural choices and domain pre-training still separate winners from also-rans.

\begin{tcolorbox}[highlightbox]
\emph{Breakthroughs appear to coincide with either massive scale increases or targeted domain pre-training, not incremental release updates.}
\end{tcolorbox}

\begin{figure}[ht]
\begin{center}
\centerline{\includegraphics[width=\columnwidth]{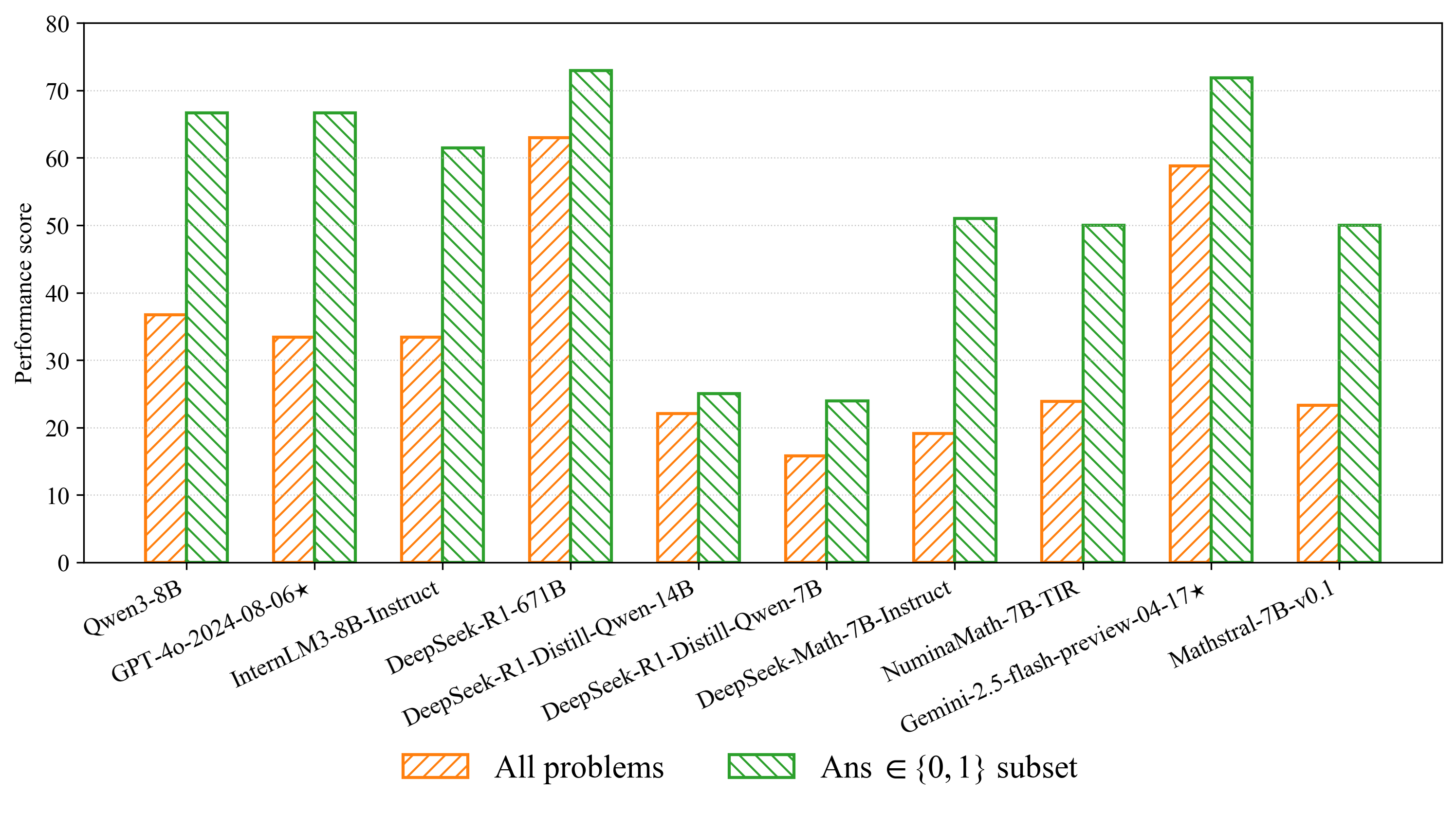}}
\caption{Pass@1 on the full RIMO-N (dark bars) versus on the 96-problem subset whose answers lie in $\{0,1\}$ (light bars). The systematic boost highlights how much easier binary outputs are for current LLMs.}
\label{fig:subset}
\end{center}
\vskip -0.3in
\end{figure}

\paragraph{Does Answer Sparsity Change the Game?}
Binary-valued items (in Olympiad terms, many of these are \emph{true-or-false} statements whose proof reduces to deciding whether the claim is correct) substantially relax the search space for an LLM.
RIMO-N contains 96 such problems whose ground-truth integer happens to be 0 or 1. Roughly two-thirds of them are genuine T/F formulations (e.g.\ ``prove that the two circumcircles have no common point''), while the remainder still ask for a numeric extremum that just evaluates to 0 or 1. Figure \ref{fig:subset} plots accuracy on this ``binary subset'' alongside full-set accuracy.
Figure \ref{fig:subset} illustrates the effect.
Scores jump by 8 to 30 percentage pointsacross in every baseline: DeepSeek-R1-671B climbs from 63\% to 73\%, Qwen3-8B from 37\% to 67\%, and the weakest system, DeepSeek-R1-Distill-Qwen-7B, rises from 16\% to 24\%. While random guessing yields 50\% on a strict T/F task, the persistent margin above chance shows that models exploit more than luck, yet the consistent gap confirms that having only two admissible outputs removes a significant portion of RIMO's challenge. In other words, part of the benchmark's hardness comes from forcing models to locate the exact integer on a larger numerical spectrum, not merely to affirm or deny a statement.

\begin{tcolorbox}[highlightbox]
\emph{A significant portion of the challenge of RIMO challenge comes from forcing models to locate an integer in a large numerical space. Restricting the answers to a binary choice substantially inflates accuracy across all models.}
\end{tcolorbox}

\begin{figure}[ht]
\begin{center}
\centerline{\includegraphics[width=1\columnwidth]{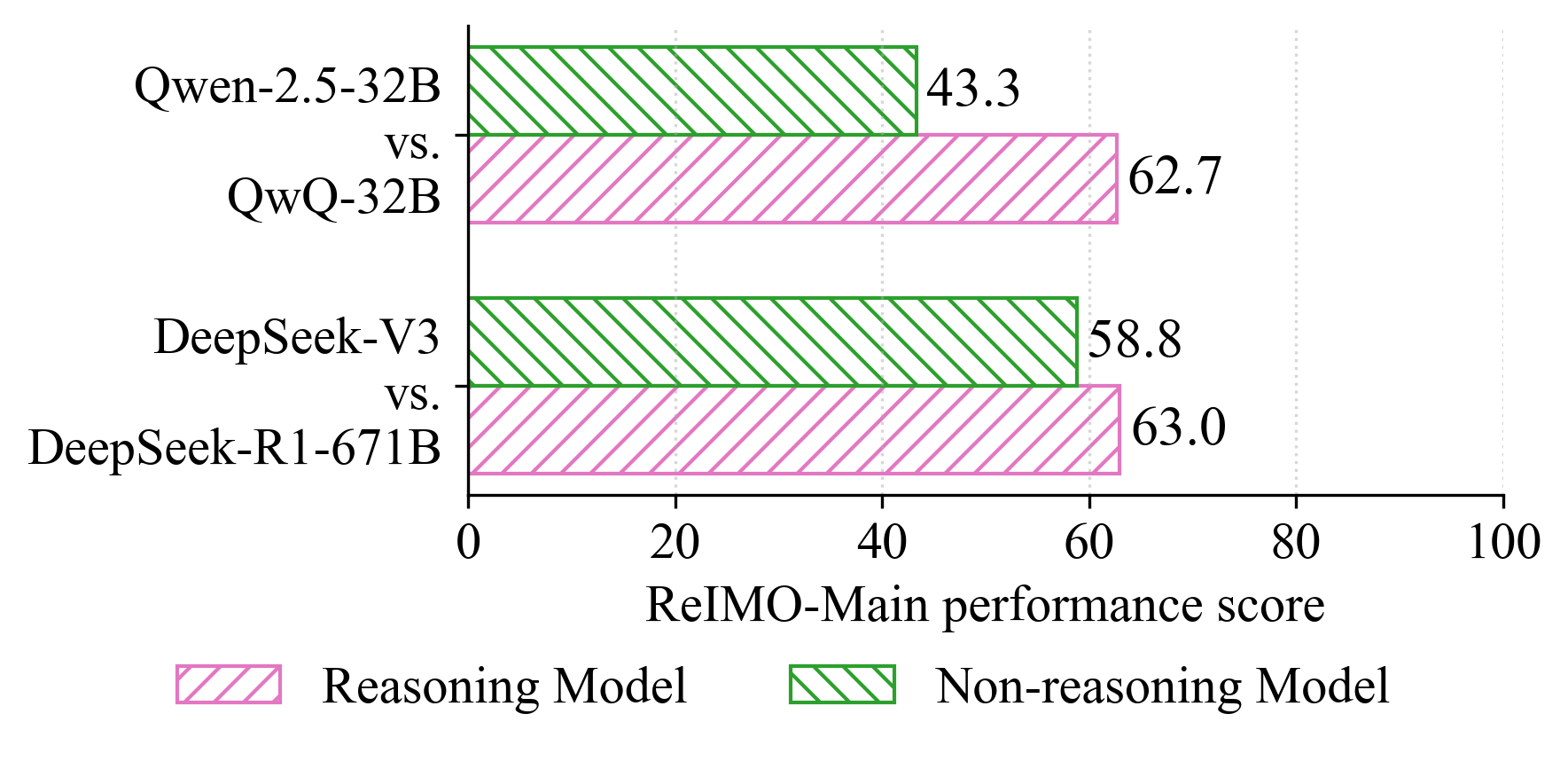}}
\caption{RIMO-N accuracy: reasoning-optimised models (dark bars) vs.\ size-matched vanilla counterparts (light bars).}
\label{fig:reasoning}
\end{center}
\vskip -0.3in
\end{figure}

\paragraph{Reasoning Model vs.\ Non-reasoning Model:}
Recent releases such as QwQ-32B and DeepSeek-R1 adopt explicit reasoning objectives (self-refinement, chain-of-thought distillation, or specialized reward modeling) on top of a backbone shipped in a ``plain'' form.  
Figure~\ref{fig:reasoning} contrasts each reasoning model with its non-reasoning sibling at an identical or near-identical scale.  
On RIMO-N the reasoning variants consistently win: QwQ-32B outperforms Qwen-2.5-32B by \textbf{19.4 percentage points} (62.7 vs.\ 43.3) and DeepSeek-R1-671B edges out the newly released DeepSeek-V3 by \textbf{4.2 percentage points} (63.0 vs.\ 58.8).  
The margin is huge when the base model is instruction-oriented but not maths-centric (Qwen-2.5).  
These results indicate that explicit reasoning optimization yields tangible gains even at the Olympiad level, over and above what scale or generic instruction tuning alone can offer.

\begin{tcolorbox}[highlightbox]
\emph{Explicit optimization for reasoning provides tangible gains at the Olympiad level, offering performance improvements over and above what scale or generic instruction tuning alone can provide.}
\end{tcolorbox}

\subsubsection{RIMO-P Results}\label{ssec:proof-results}

For \textbf{RIMO-P}, we use a sequential evaluation protocol to assess the step-by-step reasoning process. A proof attempt is graded based on the number of consecutive sub-problems solved correctly, judged by \textbf{deepseek-r1} (the top RIMO-N model). The final performance score, $P$, is the average proportion of completed steps across all problems, calculated as:
\begin{equation}\label{eq:rimo_p_score}
P = \frac{1}{N} \sum_{i=1}^{N} \frac{S_i}{X_i}
\end{equation}
where $N$ is the total number of problems (456), $S_i$ is the number of consecutively correct sub-solutions for problem $i$, and $X_i$ is its total number of sub-problems.

\begin{figure}[ht]
\begin{center}
\centerline{\includegraphics[width=1\columnwidth]{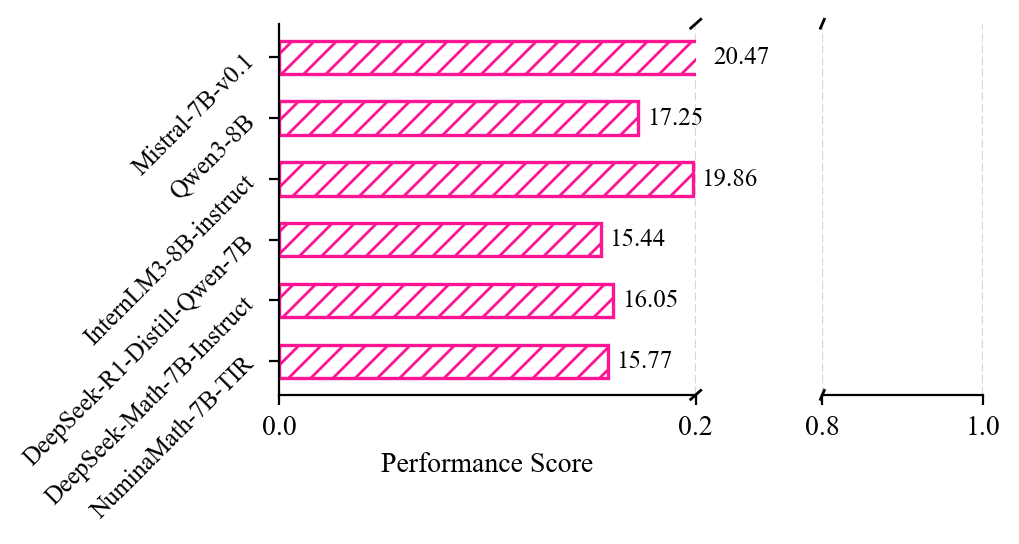}}
\caption{The performance scores of Qwen3-8B, InternLM3-8B-instruct and Mathstral-7B in RIMO-P.}
\label{fig:proof-score}
\end{center}
\vskip -0.3in
\end{figure}

Performance on this track is very low across all models (Figure~\ref{fig:proof-score}). Crucially, strong RIMO-N scores do not guarantee success on RIMO-P. This indicates that answer-finding and rigorous proof-writing are distinct capabilities, and current models struggle significantly with the latter.

From the performance, we draw three primary conclusions:
\begin{itemize}
\item \textbf{Domain-Specific Fine-Tuning Helps}: Domain-specific fine-tuning significantly aids proof generation, as shown by Mathstral's specialized LoRA pushing it ahead of equally sized generic models.
\item \textbf{Specialized Training Outweighs Recency}: Newer instruction corpora alone are insufficient for better proofs. While the recent InternLM3-8B beats Qwen, it still trails the specialized Mathstral, indicating training type is more critical than recency.
\item \textbf{A Persistent ``Proof Gap''}: All models remain far from human-level performance, leaving a large ``proof gap'' compared to advanced students. Closing this will likely require new approaches like stronger step verification, external tool use, and explicit proof planning, rather than relying on sheer scale or release cadence alone.
\end{itemize}

\section{Discussion}\label{sec:discussion}

Our results on RIMO clearly show the reasoning gap between current LLMs and Olympiad-level mathematics. The different outcomes on RIMO-N versus RIMO-P indicate that \textbf{answer-finding} and \textbf{rigorous proof-writing} are distinct skills that models currently struggle with. While our evaluation provides a clear signal of this gap, it is constrained by limited access to GPUs and the high cost of proprietary model APIs, which prevented testing of a wider range of state-of-the-art models. Even so, RIMO's noise-free framework remains a dependable tool for tracking real progress as these systems evolve.


Our analysis further suggests that performance on these challenging problems is not solely dictated by scale or recency, with the latter showing only a weak correlation of $0.21$ with accuracy. Instead, the most tangible gains come from explicit reasoning optimization, which improves performance by up to $19.4$ percentage points over vanilla counterparts. This, combined with the dramatic accuracy jump on binary-answer questions, implies that progress hinges more on targeted training objectives and problem formulation than on raw model size or release cadence.

Finally, proof generation remains a formidable frontier. The low scores on RIMO-P highlight a profound deficit in deductive ability. This represents a large gap in the performance of advanced human students, all of whom used the same rubric. Closing this gap will likely require innovative approaches beyond simple scaling, such as stronger verification loops, structured decomposition techniques, and the integration of symbolic methods.

\section{Future Work}\label{sec:future}


Future work will build on this foundation. First, we plan to translate RIMO-P into a formal language like \textbf{LEAN} to create a benchmark for machine-verifiable proofs---the highest standard of correctness. We will also continuously expand our leaderboard as more powerful models become accessible. Finally, RIMO-P's granular, decomposed structure enables detailed error analysis to pinpoint specific model weaknesses, guiding the development of more capable and reliable AI systems.


\bibliography{example_paper}
\bibliographystyle{icml2025}

\newpage
\appendix
\onecolumn

\setlength{\abovecaptionskip}{0pt} 
\setlength{\belowcaptionskip}{0pt} 
\section{Appendix}

\subsection{Prompt}

This part presents the prompt that used to eavluate the proof answer:

\begin{promptbox}{Prompt of RIMO-P Judger}
"You are an excellent mathematician and a strict judge. Your job is to evaluate the solution of a proof problem. "
"Only approve if the solution is mathematically correct, logically sound, and free of gaps or unjustified steps.\n\n"
"Sequential grading protocol (RIMO-P): We grade a proof by the number of consecutive sub-problems solved correctly. "
f"You are evaluating sub-problem {step_index} of {total_parts}. \n\n"
"You are given: (1) the original problem; (2) the official complete solution; (3) the candidate sub-solution to evaluate.\n\n"
"Instructions:\n"
"- Judge ONLY the candidate sub-solution at this step.\n"
"- Check correctness, logical validity, and consistency with the problem and official solution.\n"
"- Be strict: any error, gap, or unjustified claim => incorrect.\n"
"- Do not grade future steps.\n\n"
"Respond in STRICT JSON with keys 'verdict' and 'reason'. No extra text.\n"
"Use one of: {\"verdict\": \"correct\"} or {\"verdict\": \"incorrect\"}.\n"
"The 'reason' must be a short sentence.\n\n"
f"Problem:\n{problem}\n\n"
f"Official complete solution:\n{reference_solution}\n\n"
f"Candidate sub-solution (step {step_index}):\n{candidate_solution}\n"
\end{promptbox}

This part presents the prompt that used to evaluate the problem in Omni-Math on GPT-4o:

\begin{promptbox}{Prompt of Omni-Math Judger on Example Problem}
# CONTEXT #
I am a teacher, and I have some high-level math problems. I am tasked with evaluating the correctness of a student's answer.
Below, I am provided with a problem and a reference answer. Additionally, a student's answer is provided. My job is to assess whether the student's answer captures the same meaning as the reference answer,
even when expressed with different wording or format.
# OBJECTIVE #
I need you to judge whether the student's answer is correct given the ground truth answer.
Your tasks include:
A. Identify Mathematical or Notational Equivalence: Pay special attention to any LaTeX expressions in both answers. Confirm that the mathematical relationships, variables, and operations conveyed are equivalent.
B. Provide a Justification: Conclude with a brief explanation as to why you believe the student's output is correct or incorrect, highlighting any key differences in meaning or content.
# STYLE #
Teaching report.
# TONE #
Professional, scientific.
# AUDIENCE #
Students. Enable them to better understand whether the answer they produce is correct.
# RESPONSE: MARKDOWN REPORT #
## Student Final Answer
[Extract the student's final answer, which is enclosed in "\\boxed{}".]
## Equivalence Judgement
[Whether the student's answer share the same meaning with the reference answer. (TRUE or FALSE)]
## Justification
[Conclude with a brief explanation as to why you believe the student's answer is correct or incorrect.]
# ATTENTION #
 - The reference answer is ALWAYS correct. You should carefully judge whether the student gives the same answer as reference answer.
 - The Equivalence Judgement is only TRUE or FALSE. The answer is FALSE even if the student's final answer almost correct with a minor mistakes.
 - The answer is contained within the "boxed" section, so you can focus solely on comparing the content in the student's answer box with the reference answer, without needing to consider the intermediate steps.
 - Add "=== report over ===" at the end of the report.
<example math solution>
"solution": "The determinant equals $(-1)^{\\lceil n/2 \\rceil-1} 2 \\lceil \\frac{n}{2} \\rceil$. To begin with, we read off the following features of $S$. \\begin{itemize} \\item $S$ is symmetric: $S_{ij} = S_{ji}$ for all $i,j$, corresponding to $(a,b) \\mapsto (b,a)$). 
\\item $S_{11} = n+1$, corresponding to $(a,b) = (0,n),(1,n-1),\\dots,(n,0)$.
\\item If $n = 2m$ is even, then $S_{mj} = 3$ for $j=1,m$, corresponding to $(a,b) = (2,0),(1,\\frac{n}{2j}),(0,\\frac{n}{j})$.
\\item For $\\frac{n}{2} < i \\leq n$, $S_{ij} = \\# (\\ZZ \\cap \\{\\frac{n-i}{j}, \\frac{n}{j}\\})$, corresponding to $(a,b) = (1, \\frac{n-i}{j}), (0, \\frac{n}{j})$.
\\end{itemize} Let $T$ be the matrix obtained from $S$ by performing row and column operations as follows: for $d=2,\\dots,n-2$,  subtract $S_{nd}$ times row $n-1$ from row $d$ and subtract $S_{nd}$ times column $n-1$ from column $d$; then subtract  row $n-1$ from row $n$ and column $n-1$ from column $n$.
Evidently $T$ is again symmetric and $\\det(T) = \\det(S)$. 
Let us examine row $i$ of $T$ for $\\frac{n}{2} < i < n-1$: \\begin{align*} T_{i1} &= S_{i1} - S_{in} S_{(n-1)1} = 2-1\\cdot 2 = 0 \\\\ T_{ij} &= S_{ij} - S_{in} S_{(n-1)j} - S_{nj}S_{i(n-1)}\\\\ & = \\begin{cases} 1 & \\mbox{if $j$ divides $n-i$} \\\\ 0 & \\mbox{otherwise}.
\\end{cases} \\quad (1 < j < n-1) \\\\ T_{i(n-1)} &= S_{i(n-1)} - S_{in} S_{(n-1)(n-1)} = 0-1\\cdot0 = 0 \\\\ T_{in} &= S_{in} - S_{in} S_{(n-1)n} - S_{i(n-1)} = 1 - 1\\cdot1 - 0 = 0.
\\end{align*} Now recall (e.g., from the expansion of a determinant in minors)  if a matrix contains an entry equal to 1 which is the unique nonzero entry in either its row or its column, then we may strike out this entry (meaning striking out the row and column containing it) at the expense of multiplying the determinant by a sign. To simplify notation, we do \\emph{not} renumber rows and columns after performing this operation.
We next verify that for the matrix $T$, for $i=2,\\dots,\\lfloor \\frac{n}{2} \\rfloor$ in turn, it is valid to strike out $(i,n-i)$ and $(n-i, i)$ at the cost of multiplying the determinant by -1.
Namely, when we reach the entry $(n-i,i)$, the only other nonzero entries in this row have the form $(n-i,j)$ where $j>1$ divides $n-i$, and those entries are in previously struck columns.
We thus compute $\\det(S) = \\det(T)$ as: \\begin{gather*} (-1)^{\\lfloor n/2 \\rfloor-1} \\det \\begin{pmatrix} n+1 & -1 & 0 \\\\ -1 & 0 & 1 \\\\ 0 & 1  & 0 \\end{pmatrix} \\mbox{for $n$ odd,} \\\\ (-1)^{\\lfloor n/2 \\rfloor-1}  \\det \\begin{pmatrix} n+1 & -1 & 2 & 0 \\\\ -1 & -1 & 1 & -1 \\\\ 2 & 1 & 0 & 1 \\\\ 0 & -1 & 1 & 0 \\end{pmatrix} \\mbox{for $n$ even.} \\end{gather*} In the odd case, we can strike the last two rows and columns (creating another negation) and then conclude at once.
In the even case, the rows and columns are labeled $1, \\frac{n}{2}, n-1, n$; by adding row/column $n-1$ to row/column $\\frac{n}{2}$, we produce \\[ (-1)^{\\lfloor n/2 \\rfloor}  \\det \\begin{pmatrix} n+1 & 1 & 2 & 0 \\\\ 1 & 1 & 1 & 0 \\\\ 2 & 1 & 0 & 1 \\\\ 0 & 0 & 1 & 0 \\end{pmatrix} \\] and we can again strike the last two rows and columns (creating another negation) and then read off the result. \\n\\n\\textbf{Remark.} One can use a similar approach to compute some related determinants. For example, let $J$ be the matrix with $J_{ij} = 1$ for all $i,j$. In terms of an indeterminate $q$, define the matrix $T$ by  \\[ T_{ij} = q^{S_{ij}}. \\] We then have \\[ \\det(T-tJ) = (-1)^{\\lceil n/2 \\rceil-1} q^{2(\\tau(n)-1)} (q-1)^{n-1}f_n(q,t) \\] where $\\tau(n)$ denotes the number of divisors of $n$ and \\[ f_n(q,t) = \\begin{cases} q^{n-1}t+q^2-2t & \\mbox{for $n$ odd,} \\\\ q^{n-1}t +q^2-qt-t & \\mbox{for $n$ even.} \\end{cases} \\] Taking $t=1$ and then dividing by $(q-1)^n$, this yields a \\emph{$q$-deformation} of the original matrix $S$."
</example math solution>
## Student Final Answer
f(n)=\left\{\begin{array}{ll}
(-1)^{\frac{n}{2}-1} \cdot n & \text { if } n \text { is even } \\
(-1)^{\frac{n-1}{2}} \cdot(n+1) & \text { if } n \text { is odd }
\end{array}\right.
\end{promptbox}

\subsection{The Example of GPT-4o Failed}

\begin{explainbox}{Student Evaluation Example}
\paragraph{Student Final Answer}
\[
f(n)=
\begin{cases}
(-1)^{\frac{n}{2}-1}\,n, & n\text{ even},\\[2pt]
(-1)^{\frac{n+1}{2}}\,(n+1), & n\text{ odd}.
\end{cases}
\]

\paragraph{Equivalence Judgement}
\textbf{FALSE}

\paragraph{Justification}
The reference answer states the determinant as
\[
(-1)^{\lceil n/2\rceil-1}\,\cdot 2\cdot \lceil n/2\rceil .
\]

Break it down by the parity of $n$:
\begin{itemize}
  \item If $n$ is even, $\lceil n/2\rceil=n/2$, so
  \[
  (-1)^{n/2-1}\,2\,(n/2)=(-1)^{n/2-1}\,n .
  \]
  \item If $n$ is odd, $\lceil n/2\rceil=(n+1)/2$, so
  \[
  (-1)^{(n+1)/2-1}\,2\,(n+1)/2
  =(-1)^{(n+1)/2}\,(n+1).
  \]
\end{itemize}

Thus we again obtain
\[
f(n)=
\begin{cases}
(-1)^{\frac{n}{2}-1}\,n, & n\text{ even},\\[2pt]
(-1)^{\frac{n+1}{2}}\,(n+1), & n\text{ odd}.
\end{cases}
\]

Although mathematically identical to the student's boxed expression,
the rubric demands the \emph{exact} boxed form
\[
(-1)^{\lceil n/2\rceil-1}\,2\,\lceil n/2\rceil ,
\]
so the simplified piece-wise answer is graded as incorrect.
\end{explainbox}


\end{document}